\pdfoutput=1

\documentclass[11pt]{article}

\usepackage[preprint]{acl}

\usepackage{times}
\usepackage{latexsym}
\usepackage{soul}
\usepackage{amssymb}
\usepackage{listings}
\usepackage{multicol}  

\usepackage{tcolorbox}
\tcbuselibrary{listingsutf8}
\usepackage{xcolor}

\definecolor{sublimeBG}{HTML}{f5f5f5}
\definecolor{sublimeFG}{HTML}{393939}
\definecolor{sublimeComment}{HTML}{75715e}
\definecolor{sublimeString}{HTML}{0f1a2e}
\definecolor{sublimeKey}{HTML}{589475}
\definecolor{sublimeType}{HTML}{ae81ff}
\definecolor{sublimeNumber}{HTML}{d19a66}

\lstdefinestyle{sublime}{
  backgroundcolor=\color{sublimeBG},
  basicstyle=\ttfamily\small\color{sublimeFG},
  keywordstyle=\color{sublimeKey}\bfseries,
  stringstyle=\color{sublimeString},
  commentstyle=\color{sublimeComment}\itshape,
  numberstyle=\color{sublimeType},
  showstringspaces=false,
  breaklines=true,
  frame=none,
  columns=fullflexible,
  tabsize=1    
}

\usepackage[T1]{fontenc}

\usepackage[utf8]{inputenc}
\usepackage{enumitem}

\usepackage{microtype}

\usepackage{inconsolata}
\usepackage{amsmath}
\usepackage{algorithm}
\usepackage[noend]{algorithmic}

\usepackage{graphicx}
\usepackage{enumerate}
\usepackage{enumitem}
\usepackage{booktabs}
\usepackage{colortbl}
\usepackage[dvipsnames]{xcolor}
\definecolor{lightgray}{gray}{0.9} 
\usepackage{multirow}
\usepackage{dirtytalk}
\usepackage{listings,xcolor}
\lstdefinelanguage{json}{  
    keywords={true,false,null},  
    sensitive=true,  
    comment=[l]{//},  
    morecomment=[s]{/*}{*/},  
    string=[b]" 
}  
\usepackage{fontawesome5} 

\title{TableRAG: A Retrieval Augmented Generation Framework for Heterogeneous Document Reasoning}

\author{Xiaohan Yu$^{*}$, Pu Jian$^{*}$, Chong Chen\textsuperscript{\textcolor{black}{\faEnvelope}} \\
         Huawei Cloud BU, Beijing\\ \texttt{\{yuxiaohan5, jinapu2, chenchong55\}@huawei.com} \\
\href{https://github.com/yxh-y/TableRAG}{%
  \includegraphics[height=12pt]{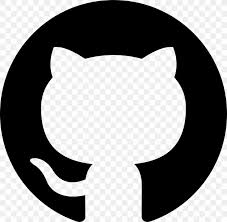}\hspace{0.3em}%
  \texttt{https://github.com/yxh-y/TableRAG}}
}

\begin{document}
\maketitle

\footnotetext[1]{$^{*}$These authors contributed equally to this work.}
\footnotetext[2] {\faEnvelope\; Corresponding author.}
\begin{abstract}
Retrieval-Augmented Generation (RAG) has demonstrated considerable effectiveness in open-domain question answering. However, when applied to heterogeneous documents, comprising both textual and tabular components, existing RAG approaches exhibit critical limitations. The prevailing practice of flattening tables and chunking strategies disrupts the intrinsic tabular structure, leads to information loss, and undermines the reasoning capabilities of LLMs in multi-hop, global queries. To address these challenges, we propose TableRAG, an SQL-based framework that unifies textual understanding and complex manipulations over tabular data. TableRAG iteratively operates in four steps: context-sensitive query decomposition, text retrieval, SQL programming and execution, and compositional intermediate answer generation. We also develop HeteQA, a novel benchmark designed to evaluate the multi-hop heterogeneous reasoning capabilities. Experimental results demonstrate that TableRAG consistently outperforms existing baselines on both public datasets and our HeteQA, establishing a new state-of-the-art for heterogeneous document question answering. 
\end{abstract}

\section{Introduction}
Heterogeneous document-based question answering \citep{chen-etal-2020-hybridqa}, which necessitates reasoning over both unstructured text and structured tabular data, presents substantial challenges.  
Tables are characterized by interdependent rows and columns, while natural language texts are sequential. Bridging this divergence within a unified QA system remains a non-trivial task.


\begin{figure}
    \centering
    \includegraphics[width=\linewidth]{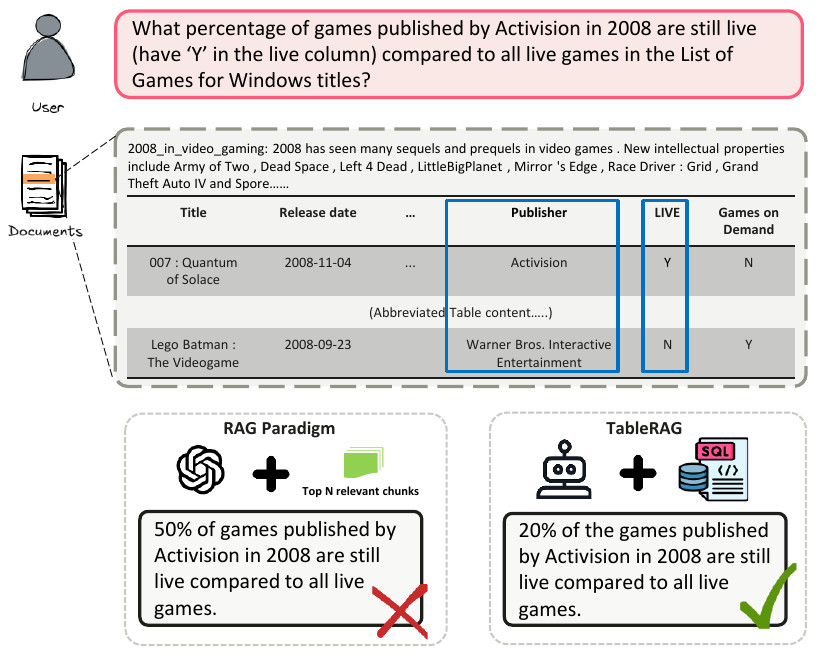}
    \caption{An example of the heterogeneous document based question answering task.}
    \label{intro}
\end{figure}

The prevailing approach extends the retrieval-augmented generation (RAG) paradigm, in which the tables are linearized into textual representations (e.g., Markdown) \citep{gao2023retrieval,jin2023tab,10.1145/3539618.3591708}. Typically, chunking strategies are employed \citep{finardi2024chronicles}, wherein flattened tables are segmented and merged with adjacent text spans. During inference, the LLMs generate answers based on the top-N retrieved chunks.
However, these methodologies are predominantly tailored to scenarios that require only surface-level comprehension of tables, such as direct answer extraction \cite{wikitq,zhu2021tat}.
When applied to extensive documents that interleave textual and tabular elements, existing RAG methodologies exhibit critical limitations:
\begin{itemize}[leftmargin=*]
    \item Structural Information Loss: The tabular structure integrity is compromised, leading to information loss or irrelevant context that impedes downstream LLMs performance.
     \item Lack of Global View: Due to document fragmentation, the RAG system struggles with multi-hop global queries \cite{edge2024local}, such as aggregation, mathematical computations, and other reasoning tasks that require a holistic understanding across entire tables.
\end{itemize}
As illustrated in Figure \ref{intro}, the RAG approach computes percentage over the top-N most relevant chunks rather than the full table, and thus results in an incorrect answer.

To address these limitations of existing RAG systems, we propose TableRAG, an SQL-based framework that dynamically transitions between textual understanding and complex manipulations over tabular data. TableRAG interacts with tables by leveraging SQL as an interface. Concretely, the framework operates via a two-stage process: an offline database construction phase and an online inference phase of iterative reasoning. The iterative reasoning procedure comprises four core operations: (i) context-sensitive query decomposition, (ii) text retrieval, (iii) SQL programming and execution, and (iv) intermediate answer generation. The utilization of SQL enables precise symbolic execution by treating table-related queries as indivisible reasoning units, thereby enhancing both computational efficiency and reasoning fidelity.
To facilitate rigorous evaluation of multi-hop reasoning over heterogeneous documents, we introduce HeteQA, a novel benchmark consisting of 304 examples across nine diverse domains. Each example contains a composition across five distinct tabular operations.
We evaluate TableRAG on both established public benchmarks and our HeteQA dataset against strong baselines, including generic RAG and program-aided approaches. Experimental results demonstrate that TableRAG consistently achieves state-of-the-art performance.
Overall, our contributions are summarized as follows:
\begin{itemize}[leftmargin=*]
    \item We identify two key limitations of existing RAG approaches in the context of heterogeneous document question answering: structural information loss and lack of global view. 
    \item We propose TableRAG, an SQL-based framework that unifies textual understanding and complex manipulations over tabular data. TableRAG comprises an offline database construction phase and a four-step online iterative reasoning process.
    \item We develop HeteQA, a benchmark for evaluating multi-hop heterogeneous reasoning capabilities. Experimental results show that TableRAG outperforms RAG and programmatic approaches on HeteQA and public benchmarks, establishing a state-of-the-art solution.
\end{itemize}





\section{Task Formulation }
In the context of the heterogeneous document question answering task, we define the task input as extensive documents, denoted as $(\mathcal{T}, \mathcal{D})$ where $\mathcal{T}$ denotes the textual contents and $\mathcal{D}$ refers to the tabular components. Given a user question $q$, the objective of this task is to optimize a function $\mathcal{F}$ that, given the combined textual and tabular context, can produce the correct answer $\mathcal{A}$:
\begin{equation}
    \mathcal{F}(\mathcal{D}, \mathcal{T}, q) \rightarrow \mathcal{A}.
\label{expert}
\end{equation}

\begin{figure*}
    \centering
    \includegraphics[width=\linewidth]{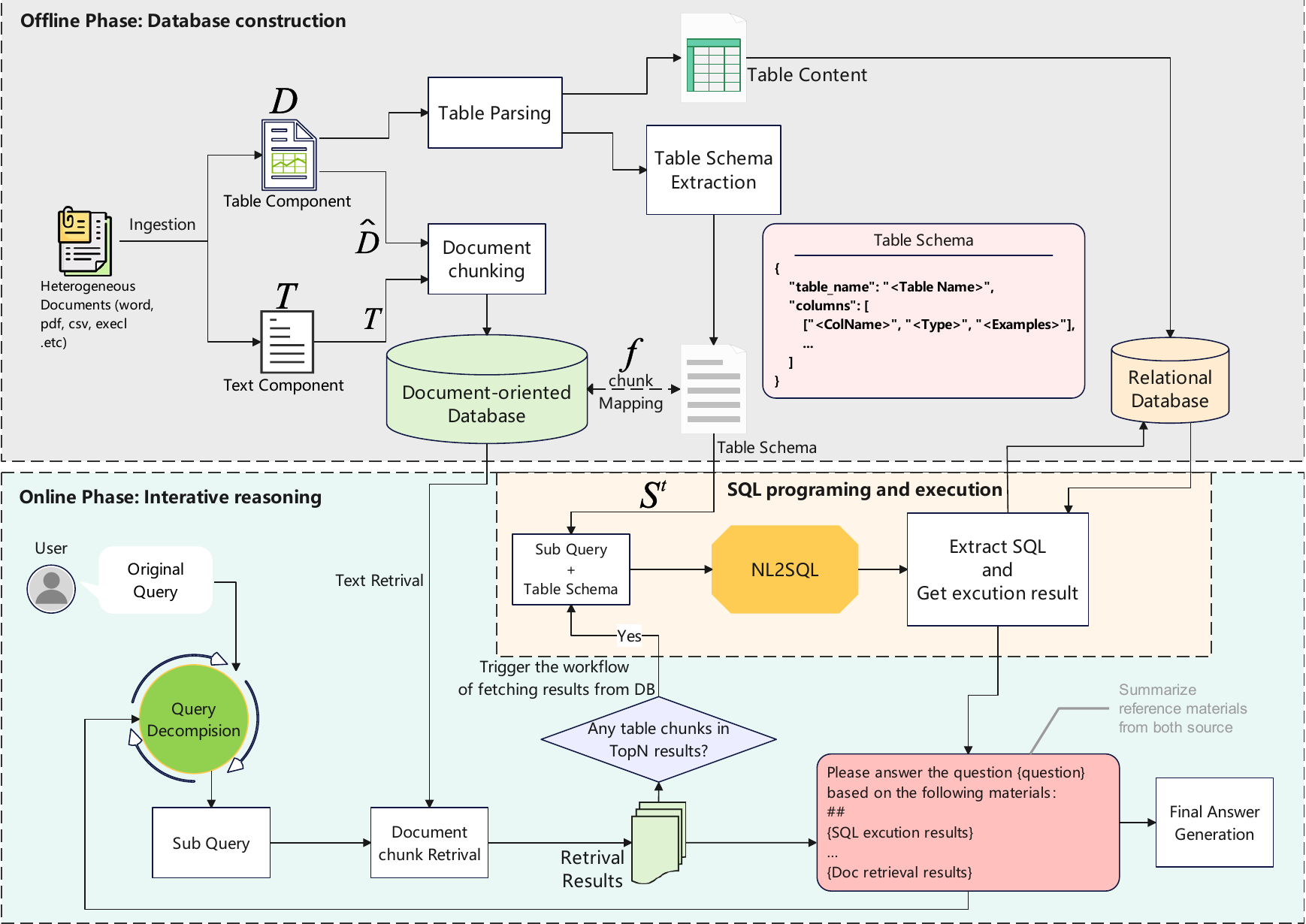}
    \caption{The overall architecture of TableRAG.}
    \label{model}
\end{figure*}

\section{TableRAG Framework}

\subsection{Overview and Design Principles}
We propose TableRAG, an SQL-based framework designed to preserve table structural integrity and facilitate heterogeneous reasoning. 
As depicted in Figure \ref{model}, TableRAG consists of offline and online workflows. The offline phase is tasked with database construction, while the online phase facilitates iterative reasoning. The reasoning procedure unfolds in a four-stage process: (i) Context-sensitive query decomposition, which identifies the respective roles of textual and tabular modalities within the query. (ii) Text retrieval. (iii) SQL programming and execution, which is selectively invoked for subqueries requiring tabular data reasoning. (iv) Compositional intermediate answer generation.
The preferential use of SQL is motivated by its capacity to leverage the expressive strength of symbolic execution over structured data, thereby enabling tabular components within user queries to be treated as monolithic reasoning units. In contrast, other languages like Python incur substantial computational overhead when dealing with large-scale data or complex workloads \citep{shahrokhi2024pytond}.

\subsection{Database Construction}
\label{database}
In the offline stage, we first extract structured components from heterogeneous documents, yielding a set of tables $\mathcal{D} = \{\mathcal{D}_1, \ldots, \mathcal{D}_M\}$. 
To enable information retrieval, we construct two parallel corpora: a textual knowledge base and a tabular schema database. 
The textual knowledge base comprises both the raw texts $\mathcal{T}$ and the Markdown-rendered form of each table, denoted as $\hat{\mathcal{D}}$. Both $\hat{\mathcal{D}}$ and $\mathcal{T}$ are segmented into chunks, which are then embedded into dense vector representations using a pre-trained language model \citep{chen2024bge}.
For tabular schema database construction, we represent each table $\mathcal{D}_i$ by a standardized schema description $S(\mathcal{D}_i)$, derived via a template as follows:
\begin{tcolorbox}[
    title={\texttt{Table Schema Template}},
    colback=sublimeBG,
    colframe=sublimeKey,
    coltitle=black,
    fonttitle=\bfseries\footnotesize,
    arc=1pt,
    boxrule=1pt,
    left=2pt,
    right=2pt,
    top=1pt,
    bottom=1pt,
]
\begin{lstlisting}[style=sublime, language=JSON]
{
    "table_name": "<Table Name>",
    "columns": [
        ["<ColName>", "<Type>", "<Examples>"],
        ...
    ]
}
\end{lstlisting}
\end{tcolorbox}
Then, we define a mapping from each flattened table chunk to its originating table schema:
\begin{align}
    f: \hat{D}_{i,j} \rightarrow S(\mathcal{D}_i)
\label{mapping}
\end{align}
where $\mathcal{\hat{D}}_{i,j}$ denotes the $j$-th chunk derived from table $\mathcal{\hat{D}}_i$. 
This mapping ensures that local segments remain contextually anchored to the table structure from which they are derived.

The tables are also ingested in a relational database (e.g., MySQL\footnote{\url{https://github.com/mysql}}), supporting symbolic query execution in the subsequent online reasoning.


\subsection{Iterative Reasoning}
To address multi-hop, global queries that require compositional reasoning over texts and tables, we introduce an iterative inference process aligned with $\mathcal{F}$ in Equation \ref{expert}. This process comprises four core operations: (i) context-sensitive query decomposition, (ii) text retrieval, (iii) program and execute SQL, and (iv) compositional intermediate answer generation. Through repeated cycles of decomposition and resolution, a solution to the query is progressively constructed. Detailed prompt templates are provided in Appendix \ref{prompts}.


\paragraph{Context-Sensitive Query Decomposition}
We explicitly delineate the respective roles of textual and tabular modalities during the reasoning process. 
While a table-related query may involve multiple semantic reasoning steps, its tabular resolution can collapse to a single executable operation. 
Consequently, an effective decomposition of global queries demands more than mere syntactic segmentation, but also structural awareness of the underlying data sources.
To this end, we first retrieve the most relevant table content from the textual database and link it to its corresponding table schema description $S(\mathcal{D}^{t})$ via the mapping function $f$. Based on this, we formulate a subquery $q_t$ at the $t$-th iteration.

\paragraph{Text Retrieval}
We deploy a retrieval module that operates in two successive stages: vector-based recall followed by semantic reranking. Given an incoming query $q_t$, it is encoded into a shared dense embedding space alongside document chunks. We then select the top-$N$ candidates with the highest cosine similarity to the query embedding:
\begin{equation}
    \mathcal{\hat{T}}^{q_t}_{recall} = \text{top-}N \left(\text{arg}\max_{\mathcal{\hat{T}}_i \in \{\mathcal{\hat{D}}, \mathcal{T}\}} cos(\mathbf{v}_{\mathcal{\hat{T}}_i}, \mathbf{v}_{q_t})\right),
\end{equation}
In the subsequent reranking stage, the recalled candidate chunks are re-evaluated by a more expressive relevance model, yielding the final top-$k$ selections, denoted by $\mathcal{\hat{T}}^{q_t}_{rerank}$.

\paragraph{SQL Programming and Execution}
To support accurate reasoning over tabular data, we incorporate a "program-and-execute" mechanism that is selectively invoked only when subquery reasoning involves tables. Specifically, we inspect whether any content originates from tabular sources in the retrieved results. For each chunk in the top-ranked set $\mathcal{\hat{T}}^{q_t}_{rerank}$, we apply the mapping function (in Equation \ref{mapping}) to extract its associated schema, yielding a table schema set:
\begin{align}
    S^t = \{f(\hat{\mathcal{T}}_i) \mid \hat{\mathcal{T}}_i \in \mathcal{\hat{T}}^{q_t}_{rerank} \}.
\end{align}
If the set $S^t$ is empty, this module is passed. Otherwise, we derive an accurate answer with the current subquery $q_t$ and the corresponding schema context as inputs.
To achieve this, we leverage structured query execution over relational data and use SQL as the intermediate formal language. A dedicated tool $f_{SQL}$ with LLM as backend generates executable SQL programs and applies them to the pre-constructed MySQL database, formalized as follows:
\begin{align}
    e_t = f_{SQL}(S^t, q_t).
\end{align}

\paragraph{Intermediate Answer Generation}
For the subquery $q_t$, TableRAG can benefit from two heterogeneous information sources: the execution result $e_t$ over SQL database and text retrieval result $\mathcal{\hat{T}}^{q_t}_{rerank}$ from the document database. Both of the data sources provide partial or complete evidence. They introduce distinct failure modes: SQL execution may produce incorrect results or execution errors, while text retrieval may yield incomplete or misleading context. Consequently, the results from these sources may either reinforce each other or present contradictions. 
To address this, we adopt a compositional reasoning mechanism. The execution result $e_t$ and the retrieved textual chunks $\mathcal{\hat{T}}^{q_t}_{rerank}$ are cross-examined to validate consistency and guide answer selection. The final answer to each subquery is derived by adaptively weighting the reliability of each source based on its evidential utility, $a_t=\mathcal{F}(e_t, \mathcal{\hat{T}}^{q_t}_{rerank})$.  

Once the query decomposition module determines that no further subqueries are necessary, TableRAG terminates the iterative reasoning process, yielding the final answer $\mathcal{A} = a_T$, where $T$ denotes the total number of iterations performed.

\section{Benchmark Construction}
In this section, we present HeteQA, a novel benchmark for assessing multi-hop reasoning across heterogeneous documents.

\subsection{Data Collection}
HeteQA necessitates advanced operations, such as arithmetic computation, nested logic, etc. To balance annotation fidelity with scalability, we adopt a human-in-the-loop collaborative strategy that integrates LLMs with human verification.  The construction pipeline proceeds in three stages:

\paragraph{Query Generation} 
We curate tabular sources from the Wikipedia dataset \citep{chen-etal-2020-hybridqa}. To facilitate analytical depth, we restrict our selection to tables with a minimum of 20 rows and 7 columns, and apply structural deduplication to eliminate redundancy across similar schemas. 
For each retained table, we define a suite of advanced operations, e.g., conditional filtering, and statistical aggregation. These operations serve as primitives for constructing complex queries. Leveraging the Claude-3.7-sonnet \footnote{\url{https://www.anthropic.com/claude/sonnet}}, we prompt for query synthesis as compositions over these primitives. Each generated query is paired with executable code in both SQL and Python.
We execute the associated code and obtain the answer. A final deduplication pass is applied over both queries and answers, promoting diversity in the dataset. Full implementation details are provided in Appendix~\ref{benchmark}.

\paragraph{Answer Verification} 
To ensure correctness and reliability, each instance is subjected to manual inspection by human annotators. Their task is to verify that the execution outcome is accurate for the corresponding query. In cases where discrepancies are found, they are responsible for correcting both the underlying code and the resulting answer.

\paragraph{Document Reference}
\label{reference}
To support queries that integrate both tabular and textual information, we augment the instance by leveraging the associated Wikipedia document. Specifically, certain entities within the query are replaced with reference-based formulations by the human annotators. For example, the query \textit{"Which driver $\ldots$"} can be rephrased as \textit{"What is the nationality of the driver $\ldots$"}. This entity substitution can either modify the subject of the question and its corresponding answer or alter the query phrasing while preserving the original answer.
The annotation guidelines and annotator profiles are detailed in Appendix \ref{instructions}.

\begin{figure}
    \centering
    \includegraphics[width=0.47\linewidth]{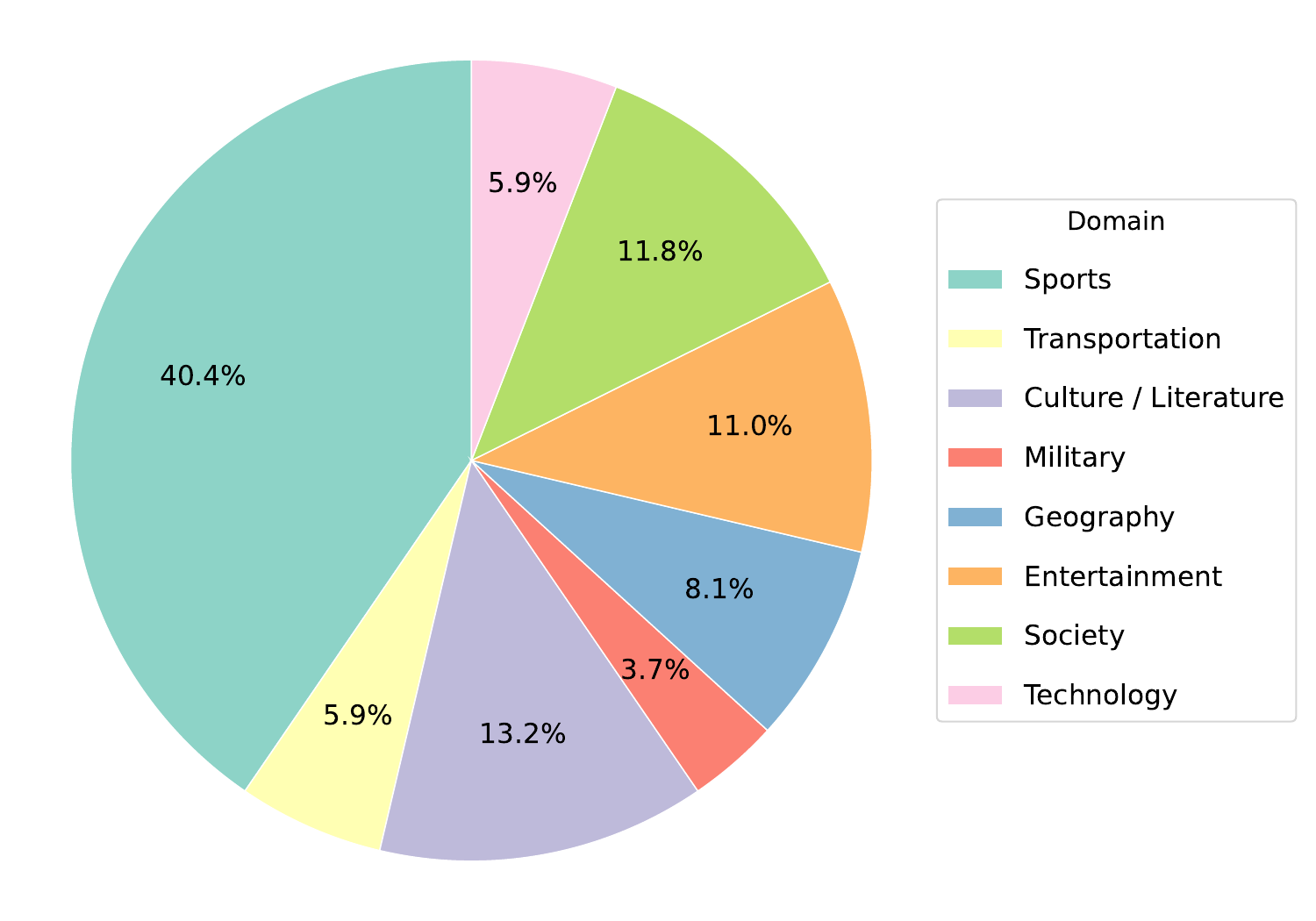}
    \includegraphics[width=0.47\linewidth]{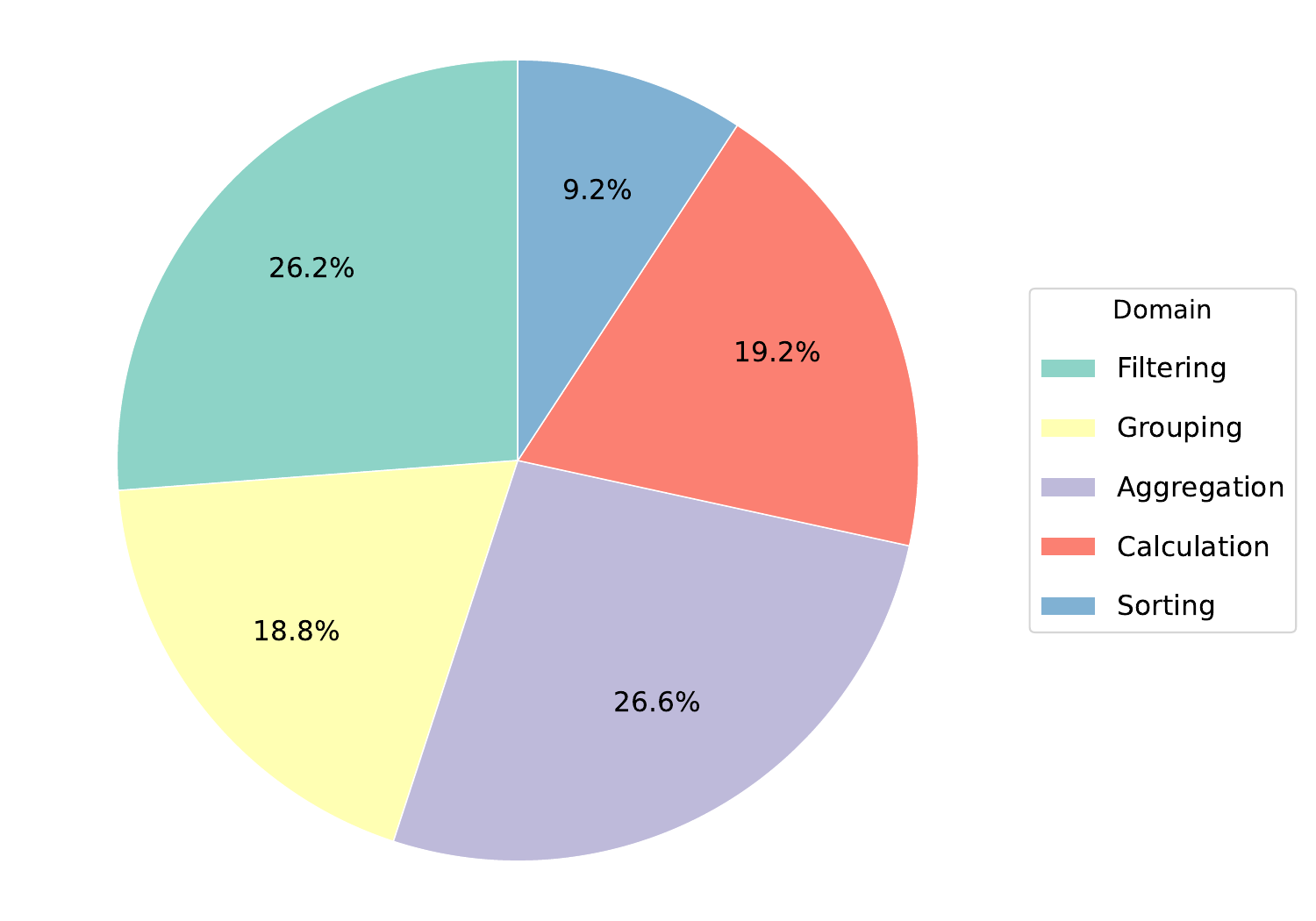}
    \caption{Domain distribution and tabular operation distribution of HeteQA.}
    \label{fig:distribution}
\end{figure}

\subsection{Discussion}
Each data instance in HeteQA is composed of a query, its corresponding answer, the executable SQL sentence, and the execution-derived answer. Through our data collection pipeline, we construct 304 high-quality examples whose answers are grounded in both single-source (82\%) and multi-source (18\%). 
The resulting benchmark spans 136 distinct tables and 5314 wiki knowledge entities.
To characterize the dataset, we analyze its semantic domains and the types of tabular reasoning operations. As illustrated in Figure~\ref{fig:distribution}, HeteQA covers 9 semantically diverse domains and encompasses 5 principal categories of tabular operations. Together, HeteQA constitutes a structurally diverse and semantically broad resource for advancing question answering over heterogeneous documents.

\section{Experiments}

\begin{table*}[!t]
\centering
\begin{tabular}{l|l|cccccccc}
\toprule[1.5pt]
\multirow{2}{*}{\textbf{Method}} & \multirow{2}{*}{\textbf{Backbone}} & \multicolumn{1}{c}{\textbf{HybridQA}} & \textbf{WikiTQ} &  \multicolumn{3}{c}{\textbf{HeteQA}} \\
\cmidrule(lr){3-3} \cmidrule(lr){4-4} \cmidrule(lr){5-7}
& & - & - & \textbf{Single-Source} & \textbf{Multi-Source} & \textbf{Overall} \\
\midrule
Direct & Claude-3.5 & 9.84 & 6.21 & 10.68 & 8.65 & 10.00 \\
& DeepSeek-R1 & 24.42 & 12.20 & 3.40 & 13.46 & 6.77 \\
& DeepSeek-V3 & 14.75 & 10.39 & 6.80 & 28.85 & 14.19 \\
& Qwen-2.5-72b & 11.47 & 7.37 & 4.85 & 12.50 & 7.42 \\
\midrule
NaiveRAG & Claude-3.5 & 20.28 & 82.60 & 33.20 & 40.35 & 34.54 \\
& DeepSeek-V3 & 26.56 & 75.40 & 33.60 & 45.61 & 35.85 \\
& Qwen-2.5-72b & 22.62 & 66.33 & 23.07 & 36.84 & 25.66 \\
\midrule
ReAct & Claude-3.5 & 43.38 & 69.81 & 26.40 & 44.44 & 29.60 \\
& DeepSeek-V3 & 38.36 & 63.40 & 21.14 & 47.39 & 26.07 \\
& Qwen-2.5-72b & 37.38 & 53.80 & 16.94 & 35.71 & 20.47 \\
\midrule
\midrule
TableGPT2 &  & 9.51 & 63.40 & 35.60 & 16.67 & 32.24 \\
\midrule
\midrule
TableRAG & Claude-3.5 & 47.87 & \textbf{84.62} & \textbf{44.94} & 40.74 & 44.19  \\
& DeepSeek-V3 & 47.87 &	80.40 & 43.32 & \textbf{51.85} & \textbf{44.85} \\
& Qwen-2.5-72b & \textbf{48.52} & 78.00 & 37.65 & 43.96 & 38.82 \\
\bottomrule[1.5pt]
\end{tabular}
\caption{Performance of TableRAG compared to baseline models across multiple benchmarks, evaluated using LLM-as-Judge accuracy. "Multi-Source" indicates questions requiring both tabular and textual information, while “Single-Source” refers to questions relying on only one source type.}
\label{main}
\end{table*}

\subsection{Experimental Settings}

\subsubsection{Datasets.}
We assess the performance of TableRAG on our curated HeteQA, as well as multiple established benchmarks spanning two settings:

\paragraph{HybridQA}\citep{chen-etal-2020-hybridqa}
A multi-hop QA dataset involving both tabular and textual information. For our evaluation, we only retain data cases with tables containing more than 100 cells. 

\paragraph{WikiTableQuestion}\citep{pasupat2015compositional} A TableQA dataset spanning diverse domains. The queries necessitate a range of data manipulation operations, including comparison, aggregation, etc.


\subsubsection{Implementation Details}
In the text retrieval process, we employ the BGE-M3 series models \citep{chen2024bge,bge-m3}. During recall, we retain the top 30 candidates, from which the top 3 are subsequently selected via reranking. To manage large inputs, the text is chunked into segments of 1000 tokens, with a 200-token overlap between consecutive chunks. The iterative loop is bounded by a maximum of 5 iterations.
For backbone LLMs, we utilize Claude-3.5-Sonnet as a representative closed-source LLM, while Deepseek-V3, Deepseek-R1 \citep{guo2025deepseek}, and Qwen-2.5-72B \citep{yang2024qwen2} serve as the open-source counterparts. A consistent backend is maintained for all modules in the online iterative reasoning process.
We use accuracy as the evaluation metric, assessed by Qwen-2.5-72B, which yields a binary score of 0 or 1. The prompt is provided in Appendix \ref{prompts}, and the results evaluated using the exact match metric are presented in Appendix \ref{append_exp}.

\subsubsection{Baselines}
We evaluate the performance of TableRAG by benchmarking it against three distinct baseline methodologies: (1) Direct answer generation with LLMs. (2) NaiveRAG, which processes tabular data as linearized Markdown formatted texts and subsequently applies a standard RAG pipeline. (3) ReAct \citep{yao2023react}, a prompt based paradigm to synergize reasoning and acting in LLMs with external knowledge sources. (4) TableGPT2 \citep{su2024tablegpt2} employs a Python-based execution module to generate code (e.g., \texttt{Pandas}) for answer derivation within a simulated environment.
The detailed implementation of these baseline methods is provided in Appendix \ref{baseline} and additional baseline comparisons are presented in Appendix \ref{append_exp}.

\subsection{Main Result}
The main results across different LLMs backbones are presented in Table \ref{main}. Several key observations emerge: (1) The ReAct framework demonstrates advantages over naive RAG on multi-source data, but exhibits degraded performance on single-source data that requires tabular reasoning. This can be attributed to context insensitivity during multi-turn reasoning. Queries that could be resolved with single SQL execution are instead decomposed into multiple subqueries. This over-fragmentation may introduce errors or incomplete information during execution (especially when operations such as filtering, aggregation are involved), potentially leading to cascading failures.
(2) TableGPT2 yields acceptable results solely on single-source queries, such as WikiTQ, underscoring its limited capacity for handling multi-source queries. This reflects a lack of generalizability in heterogeneous information environments. (3) TableRAG surpasses all baselines, achieving at least a 10\% improvement over the strongest alternative. Notably, it performs robustly across both single-source and multi-source data. This performance gain is attributed to the incorporation of symbolic reasoning, which enables effective adaptation to heterogeneous documents. Moreover, the consistency in performance across different LLM backbones underscores TableRAG’s architectural generality and compatibility with a broad range of backbones.


\begin{figure}
    \centering
    \includegraphics[width=0.48\linewidth]{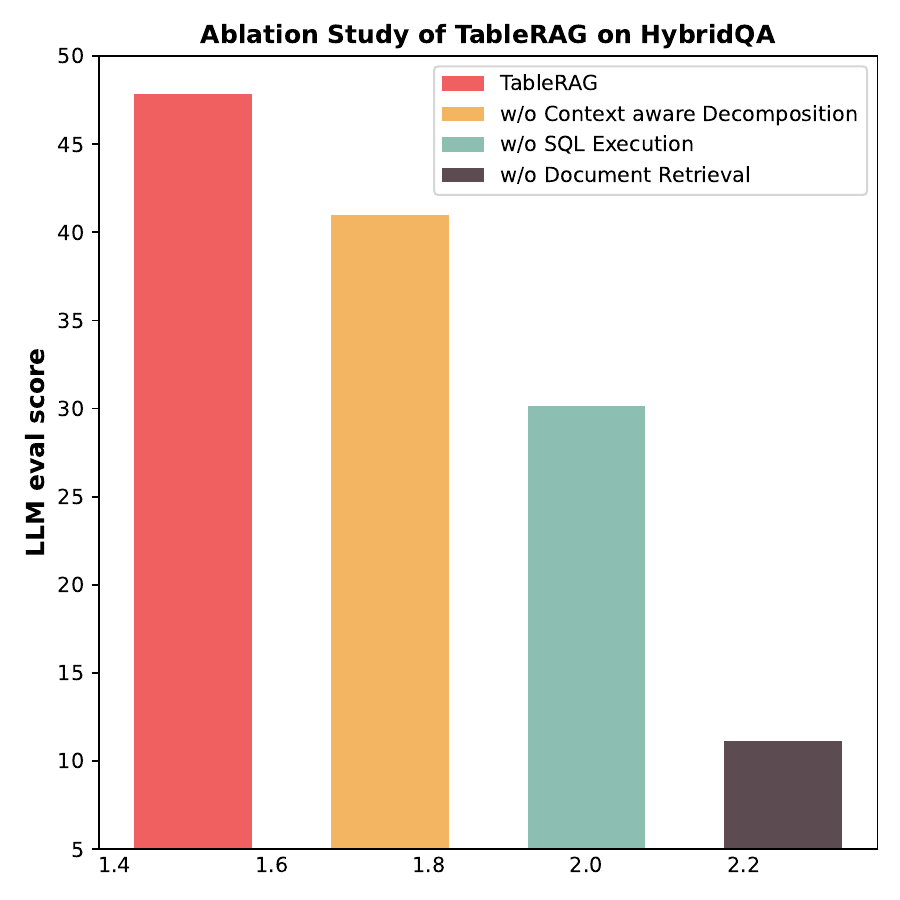}
    \includegraphics[width=0.48\linewidth]{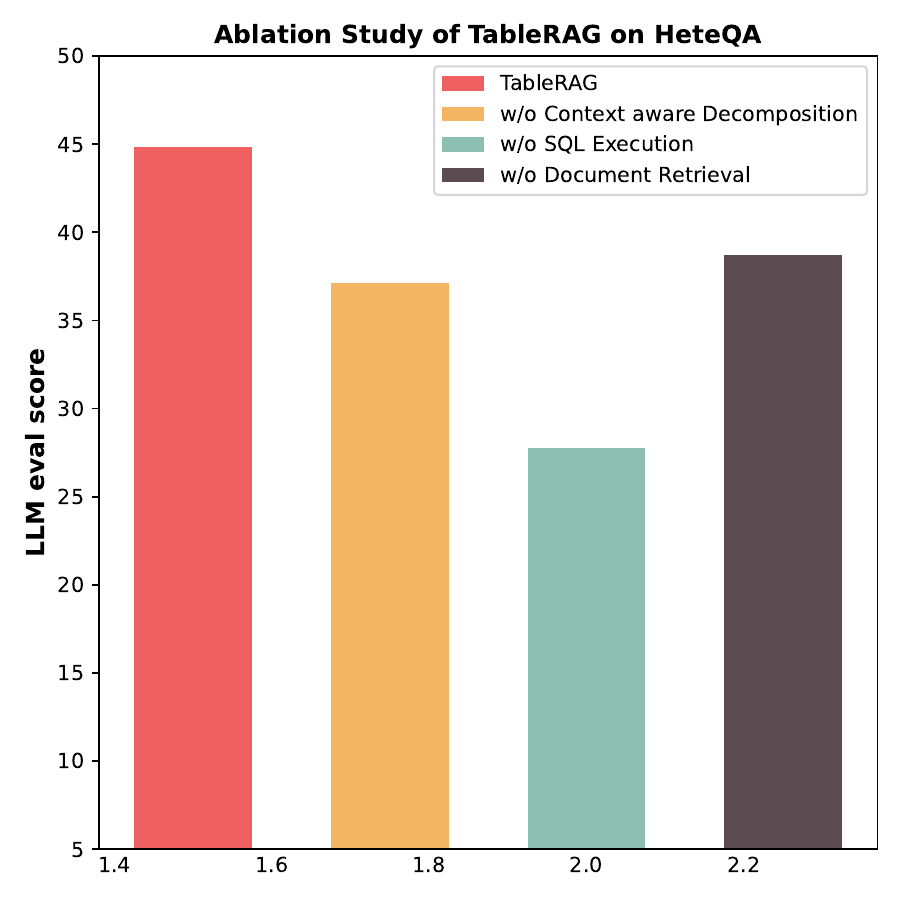}
    \caption{Ablation study on HybridQA and HeteQA benchmarks based on DeepSeek-V3 backbone.}
    \label{ab}
\end{figure}

\begin{table*}[!t]
\centering
\begin{tabular}{l|cc|cc}
\toprule[1.5pt]
\textbf{Method} & \multicolumn{2}{c|}{\textbf{WikiTQ}} & \multicolumn{2}{c}{\textbf{HeteQA}} \\ \cmidrule{2-3} \cmidrule{4-5}
& \textbf{Total Latency} & \textbf{Avg. Latency / Step} & \textbf{Total Latency} & \textbf{Avg. Latency / Step} \\
\midrule
ReAct & 13.50 & 5.92 & 24.57 & 7.70 \\
TableRAG & 17.70 & 7.93 & 45.65 & 10.79 \\
\bottomrule[1.5pt]
\end{tabular}
\caption{Latency evaluation, measured in seconds.}
\label{latency}
\end{table*}

\subsection{Ablation Study}
To elucidate the relative importance of each component within the TableRAG framework, we evaluate the full architecture against three ablated variants: (1) w/o Context-Sensitive Query Decomposition, where query decomposition is performed without conditioning on retrieved table schema. (2) w/o SQL Execution, which replaces the SQL programming and execution module with the markdown table format. (3) w/o Textual Retrieval, which operates solely through table-based SQL execution, without leveraging textual resources such as Wikipedia documents.
The results are summarized in Figure~\ref{ab}. All the modules contribute to the overall performance of TableRAG, though their relative impact varies across benchmarks. On HybridQA, document retrieval proves particularly critical, due to its emphasis on the extraction of entity-centric or numerical cues. Conversely, for HeteQA, SQL execution proves more influential, as the queries involve nested operations that benefit from SQL-based symbolic reasoning. These findings highlight the complementary design of TableRAG’s textual retrieval and program-executed reasoning components.


\begin{figure}
    \centering
    \includegraphics[width=\linewidth]{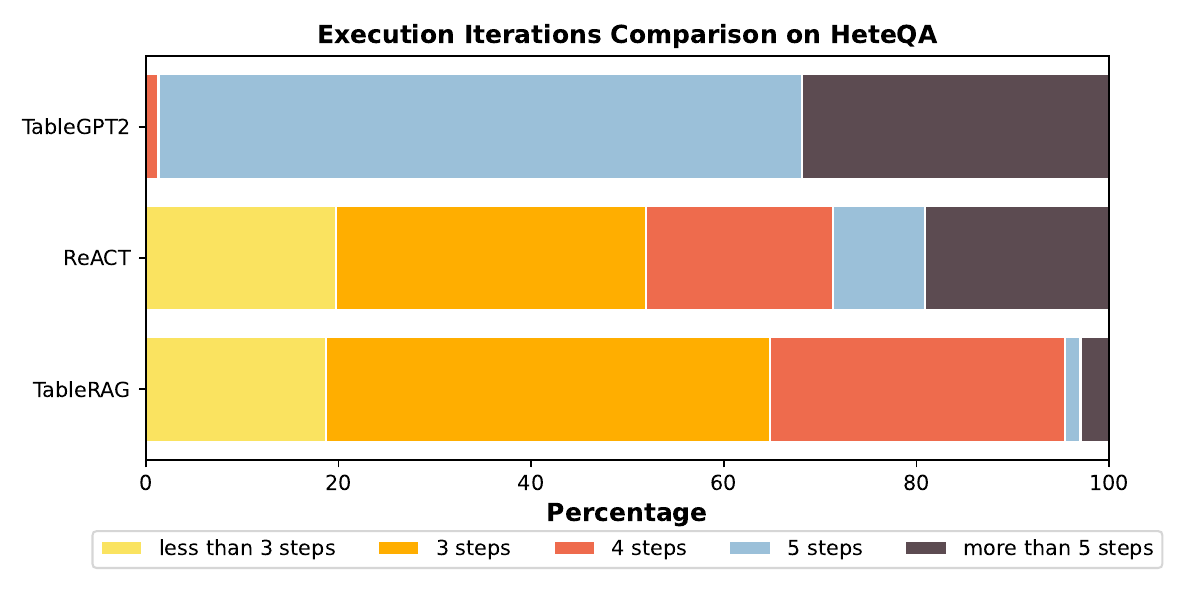}
    \caption{Comparison of the execution iterations on HeteQA between TableRAG, ReAct and TableGPT2.}
    \label{steps}
\end{figure}

\section{Efficiency}
\subsection{Execution Dynamics}
We evaluate the efficiency of TableRAG by examining the distribution of its execution iterations, as illustrated in Figure \ref{steps}. Execution lengths are grouped into four categories: fewer than 3 steps, 3–5 steps, exactly 5 steps, and more than 5 steps. 
Among the evaluated methods, TableGPT2 demonstrates the highest average number of execution steps, with a modal value centered around five. In contrast, TableRAG consistently requires fewer steps, resolving approximately 63.55\% of instances in fewer than five steps and an additional 30.00\% precisely within five, with only a marginal proportion of cases remaining unsolved under the given iteration constraints. While ReAct exhibits a comparable distribution in execution steps, its overall performance remains markedly inferior to that of TableRAG. These results suggest that TableRAG achieves both superior efficiency in execution and outstanding reasoning accuracy. It is attributed to the incorporation of SQL-based tabular reasoning.

\subsection{Latency}
We assess the latency performance using Qwen-72B-Instruct as a consistent backbone model. As shown in Table \ref{latency}, TableRAG incurs lower latency compared to ReAct, in both average time per step and total time efficiency. This is mainly due to: (1) TableRAG executes fewer iterations, and employs symbolic SQL execution, which is computationally efficient. (2) ReAct performs reasoning at every step, and consumes more steps before reaching a final decision, thereby incurring greater time overhead.

\section{Analysis}
We provide a comprehensive analysis of TableRAG in this section, with additional results presented in Appendix~\ref{analysis}.

\subsection{Error Analysis}
\begin{figure}
    \centering
    \includegraphics[width=\linewidth]{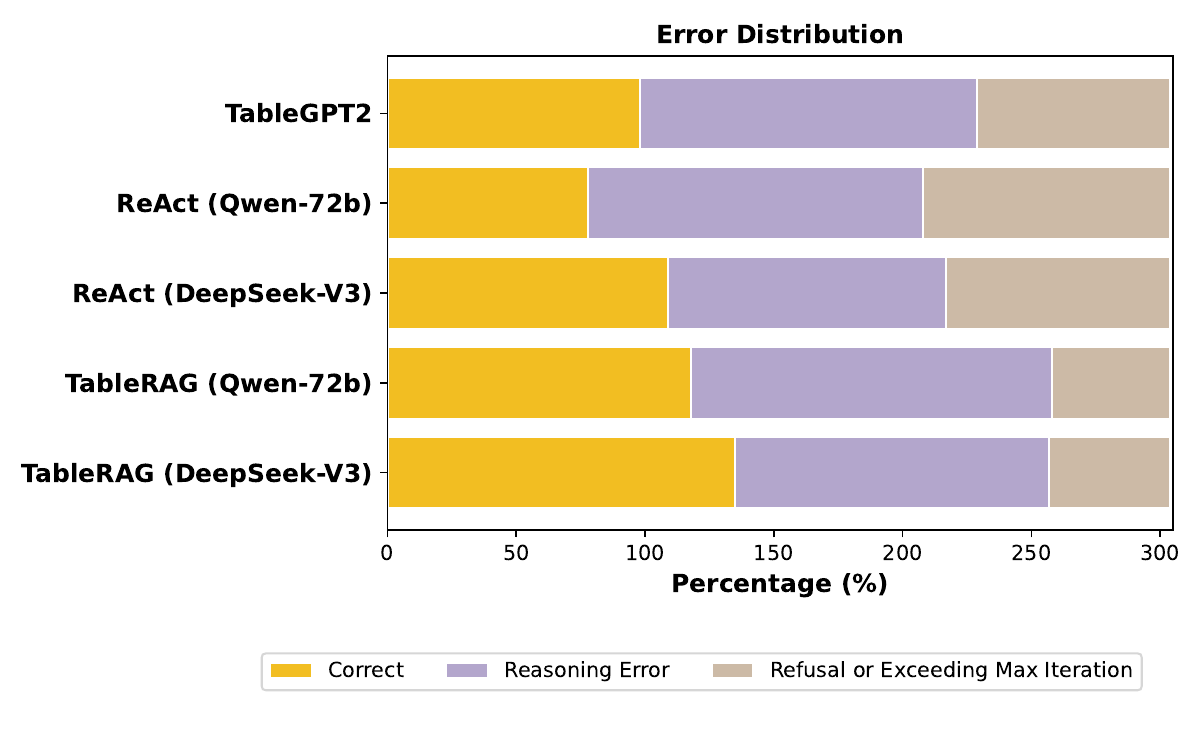}
    \caption{Error analysis of TableRAG, TableGPT2 and ReAct with DeepSeek-V3 and Qwen-2.5-72b as backbones on HeteQA.}
    \label{fig:failure}
\end{figure}

In addition to evaluating the overall performance of TableRAG against established baselines, we performed a detailed error analysis to characterize the nature of prediction failures. Broadly, the incorrect outputs fall into two primary categories: (1) reasoning failures, attributable to errors in SQL execution or flawed intermediate query decomposition, and (2) task incompletion, typically manifesting as refusals to answer or termination upon exceeding the maximum iteration limit. The prediction distribution is shown in Figure \ref{fig:failure}. Notably, TableGPT2 exhibits the highest frequency of such failures, largely due to its limited capacity to integrate contextual cues from the wiki documents. This constraint frequently results in the model either explicitly refusing to respond or acknowledging its inability to do so. In contrast, ReAct, which lacks mechanisms for context-aware query decomposition and code execution simulation, often engages in unnecessarily elaborate reasoning steps for problems that could be addressed via a single structured inquiry. TableRAG demonstrates the lowest failure rate among the methodologies assessed. Its consistent ability to yield valid responses within five iterations highlights the efficacy of its design — particularly its use of context-aware query decomposition and selective SQL-based execution planning.

\subsection{Prediction across Domains}
\begin{figure}
\centering
\includegraphics[width=0.6\linewidth]{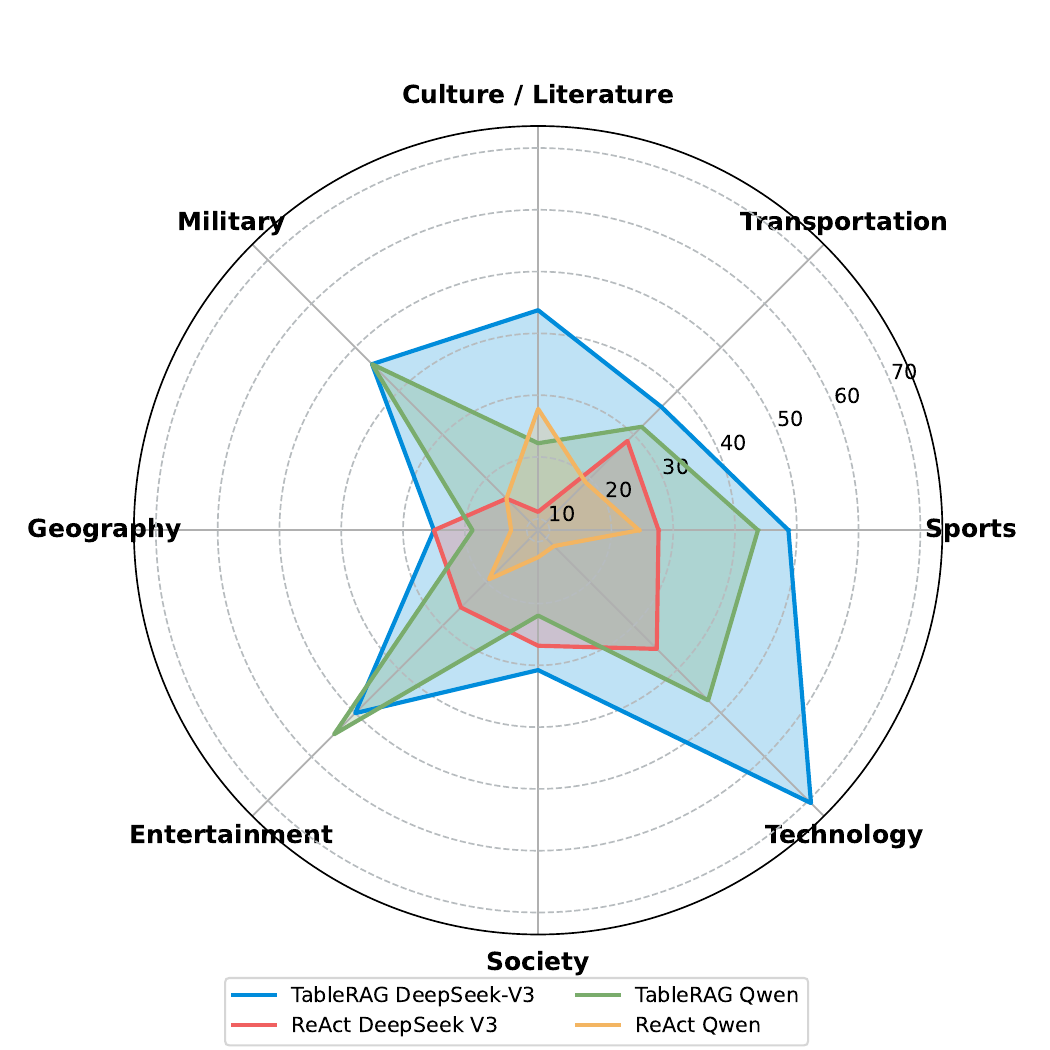}
\caption{Performance distribution of TableRAG and ReAct across different domains.}
\label{perform_domain}
\end{figure}

Figure \ref{perform_domain} presents a comparative evaluation of TableRAG, instantiated with various backbone LLMs, against the ReAct framework across various domains. The results reveal that TableRAG consistently outperforms ReAct in the majority of domains, demonstrating its effectiveness in heterogeneous document question answering. Only certain domains, such as Culture, exhibit comparatively weaker performance on TableRAG with Qwen backbone. A closer inspection of the data distribution suggests that this degradation may stem from the sparsity of domain-specific instances.

\section{Related Work}

\subsection{Retrieval Augmented Generation}
Retrieval-Augmented Generation (RAG) has emerged as a robust paradigm for mitigating hallucination \citep{zhang2023siren} and enhancing the reliability of Large Language Models (LLMs) generated responses \citep{lewis2020retrieval, guu2020retrieval}. The RAG approaches retrieve from the knowledge base, and the most relevant document chunks are subsequently incorporated into the generation process \citep{gao2023retrieval,zhu2024realm,borgeaud2022improving}. However, this straightforward retrieval process often yields noisy chunks that may lack critical details, thereby diminishing the quality of the subsequent generation. Recent advancements have thus focused on task-adaptive retrieval mechanisms. Notable frameworks in this regard include Self-RAG \citep{asai2023self}, RQ-RAG \citep{chan2024rq}, etc. Despite these innovations, RAG still faces challenges when dealing with heterogeneous contexts \citep{satpute2024can}.

\subsection{Table Reasoning via Large Language Models}
Table reasoning refers to the development of a system that provides responses to user queries based on tabular data \citep{lu2025large}. The mainstream approaches to table reasoning can be broadly classified into two categories. The first category revolves around leveraging LLMs through prompt engineering. For instance, Tab-CoT \citep{jin2023tab} applies chain-of-thought (CoT) reasoning to establish a tabular structured reasoning process. Similarly, Chain-of-Table \citep{wang2024chain} extends the CoT methodology to the tabular setting, enabling a multi-step reasoning process for more complex table-based queries. The second category involves utilizing programs to process tabular data. Tabsqlify \citep{nahid2024tabsqlify} employs a text-to-SQL approach to decompose tables into smaller, contextually relevant sub-tables. DATER \citep{ye2023large} adopts a few-shot prompting strategy to reduce large tables into more manageable sub-tables, using a parsing-execution-filling technique that generates intermediate SQL queries. BINDER \citep{cheng2022binding,zhang2023reactable} integrates both Python and SQL code to derive answers from tables. InfiAgent-DABench \citep{hu2024infiagent} utilizes an LLM-based agent that plans, writes code, interacts with a Python sandbox, and synthesizes results to solve table-based questions.

\section{Conclusion}
We address the limitations of existing RAG approaches in handling heterogeneous documents that combine textual and tabular data. Current approaches compromise the structural integrity of tables, resulting in information loss and degraded performance in global, multi-hop reasoning tasks. To overcome these issues, we introduce TableRAG, an SQL-driven framework that integrates textual understanding with precise tabular manipulation. To rigorously assess the capabilities of our approach, we also present a new benchmark HeteQA. Experimental evaluations across public datasets and HeteQA reveal that TableRAG significantly outperforms existing baseline approaches.

\section*{Limitations}
While TableRAG demonstrates strong performance, several limitations merit consideration:
1. The effectiveness of TableRAG is closely tied to the capabilities of the underlying LLMs. Our implementation leverages high-capacity models such as Claude, DeepSeek-v3, and Qwen-72B-Instruct, which possess strong generalization abilities. Smaller models that lack specialized instruction tuning may exhibit a marked degradation in performance.
This suggests that achieving competitive results may necessitate substantial computational resources.
2. The HeteQA benchmark is restricted to English. This limitation arises from the difficulty in curating high-quality heterogeneous sources across multiple languages. As a result, cross-lingual generalization remains unexplored. In future work, we aim to extend HeteQA to a multilingual setting, thereby broadening the applicability and robustness of our evaluation framework.


\bibliography{custom}

\appendix

\section{HeteQA}
\label{benchmark}

\lstset{  
    language=SQL,  
    basicstyle=\ttfamily\small,  
    keywordstyle=\color{blue}\bfseries,  
    stringstyle=\color{red},  
    commentstyle=\color{green!50!black},  
    showspaces=false,  
    showstringspaces=false,  
    showtabs=false,  
    tabsize=4,  
    morekeywords={  
        CREATE, ALTER, DROP, INSERT,   
        UPDATE, DELETE, SELECT,   
        FROM, WHERE, JOIN, GROUP BY,   
        ORDER BY, INNER, LEFT, RIGHT  
    }  
} 
\begin{table*}
\centering
\begin{tabular}{p{2cm} p{14cm}}
\toprule
\textbf{table} & List\_of\_Australian\_films\_of\_2012\_0 \\
\midrule
\textbf{query} & Who wrote and starred the comedy film released in the second half of 2012 (July-December) that had the highest number of cast members in the List of Australian films of 2012? \\
\midrule
\textbf{sql\_query} & Which comedy film released in the second half of 2012 (July-December) had the highest number of cast members in the List of Australian films of 2012? \\
\midrule
\textbf{sql} & 
\begin{lstlisting}[language=SQL]
SELECT 
  title, 
  LENGTH(cast_subject_of_documentary) - LENGTH(
    REPLACE(
      cast_subject_of_documentary, ',', 
      ''
    )
  ) + 1 AS cast_count 
FROM 
  `list_of_australian_films_of_2012_0_sheet1` 
WHERE 
  genre LIKE '%Comedy%' 
  AND (
    release_date LIKE '%July%' 
    OR release_date LIKE '%August%' 
    OR release_date LIKE '%September%' 
    OR release_date LIKE '%October%' 
    OR release_date LIKE '%November%' 
    OR release_date LIKE '\ufffdcember%'
  ) 
ORDER BY 
  cast_count DESC 
LIMIT 
  1;
\end{lstlisting} \\
\midrule
\textbf{sql\_ans} & Kath \& Kimderella \\
\midrule
\textbf{answer} & Riley, Turner, and Magda Szubanski \\
\bottomrule
\end{tabular}
\label{data_example}
\caption{An example of HeteQA.}
\end{table*}

\subsection{Table Collection}
To enable complex reasoning over tabular structures, we curate a subset of extensive tables from the Wikipedia-based corpus. Specifically, we retain only tables containing more than 20 rows and at least 7 columns. This filtering process reduced the initial collection from 15,314 tables to 1,345 candidates that meet the criteria for structural richness. 
To further enhance query diversity and reduce redundancy in the dataset, we apply a deduplication step based on schema similarity. In cases where multiple tables share identical column structures — for example, the entries for \texttt{1947 BAA draft} and \texttt{1949 BAA draft} — only a single representative instance is preserved. This procedure yielded a final dataset comprising 155 unique tables.

\begin{figure}
    \centering
    \includegraphics[width=\linewidth]{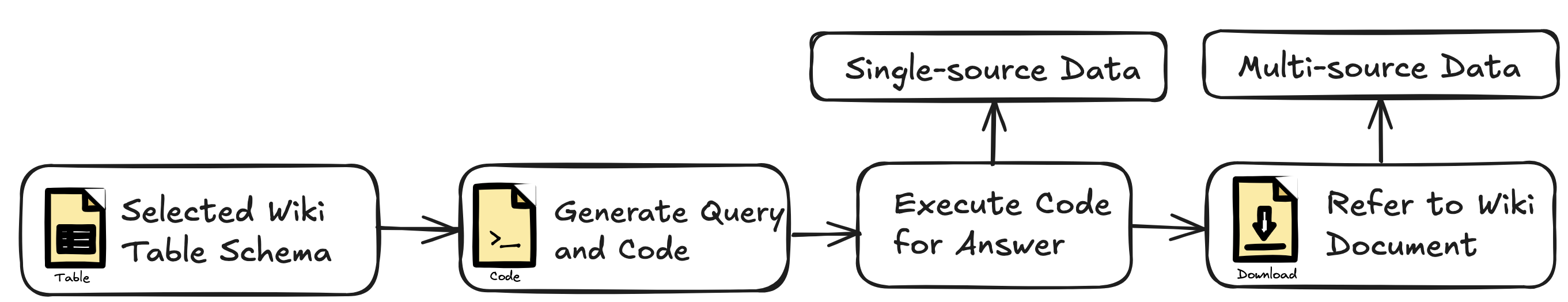}
    \caption{The dataset construction pipeline of HeteQA.}
    \label{fig:benchmark}
\end{figure}

\subsection{Data Collection}
As illustrated in Figure~\ref{fig:benchmark}, each data instance is constructed through a three-stage pipeline, facilitated via the Claude 3.7 Sonnet API\footnote{\url{https://www.anthropic.com/claude/sonnet}}.

To construct the dataset, we prompt the LLM to generate SQL or \texttt{pandas} code associated with queries conditioned on a provided table schema. The generated code is then executed against the table to obtain the corresponding answer. Instances that fail to produce an executable result are discarded.
To promote diversity within the dataset and avoid redundancy, we eliminate any instances exhibiting duplicate queries or identical answers. This filtering is performed using regular expressions to detect lexical or semantic repetition. Additionally, we discard cases where the execution result is ambiguous, such as queries seeking the top-ranked entity when multiple candidates are tied.
Following automated filtering, all remaining instances undergo a manual verification process. Two human annotators independently assess each query, the corresponding code, and the resulting answer. In cases of inconsistency or error, annotators revise the SQL or Python code and correct the associated answer to ensure accuracy and coherence.


\section{Guidelines for HeteQA Annotations}
\label{instructions}

\subsection{Annotator Profiles}
The two annotators are volunteers whose native language is Chinese and who possess fluent English proficiency. Both annotators are professional software engineers with substantial experience in SQL and Python programming. 

\subsection{Guidelines for Answer Verification}
\textbf{Objective }

\noindent Your task is to verify whether the provided answer correctly and completely addresses the natural language query, based on the associated code and table. You will also identify and handle ambiguous or non-unique cases.
\\

\noindent \textbf{Annotation Steps}

\noindent 1. Verify the Answer

\noindent Read the natural language query and examine the accompanying Excel table, SQL statements, and Python (pandas) code. Verify whether the output aligns with the semantics of the original query and the underlying data. To this end, you are permitted to execute multiple operations on the provided Excel files.

\noindent 2. Correction Tasks (if necessary)

\noindent If the answer is incorrect, modify both the code and the answer so that they correctly fulfill the query. Ensure the modified code is minimal, clean, and logically sound.

\noindent 3. Ambiguity and Tie Cases

\noindent In instances where the query yields no unique resolution - whether due to tied outcomes, under specified conditions, or semantic ambiguity in the formulation - you can employ principled strategies to ensure meaningful processing:

\noindent - Option A: Modify the query to resolve the ambiguity (e.g., make clarification).

\noindent - Option B: If the ambiguity is irreparable or the answer depends on arbitrary choices, discard the case. \\

\noindent \textbf{Additional Guidelines}

\noindent Maintain consistency between the query, code, and answer. Ensure your corrections do not introduce new ambiguities or assumptions not grounded in the table or query.

\subsection{Guidelines for Document Reference}
\textbf{Objective}

\noindent Your task is to add an additional reasoning hop to the original query using the provided Wikipedia entities and content. This will help transform the original query into a more complex, multi-hop question that requires deeper reasoning. \\

\noindent \textbf{Input Data}

\noindent Each data case consists of:

\begin{itemize}[leftmargin=*]
    \item Original Query
    \item Answer
    \item A set of Wikipedia entities relevant to the query or the answer
    \item Corresponding Wikipedia content for each entity
\end{itemize}

\noindent \textbf{Task Overview}

\noindent For each data case, check if the original answer is mentioned in the provided Wiki entities:

\noindent - If not mentioned or ambiguous (e.g., "Jordan" for both basketball star and sports brand), do not change the query and answer. Just mark the case as “No modification needed”.

\noindent - If the answer entity exists in the Wikipedia content, perform the following steps:

1. Identify key factual descriptions about the answer entity from its Wiki content.

2. Add a reasoning hop to the original query that leads to the answer via this key fact.

3. Generate two candidate (Query, Answer) pairs by rewriting the query to incorporate this extra reasoning step and updating the answer accordingly. \\

\noindent \textbf{Example}

\noindent Original Query: Which album takes first place on the Billboard leaderboard in 2013?

\noindent Original Answer: ArtPop

\noindent Wiki Entity: ArtPop (album)

\noindent A key description: It was released on November 6, 2013, by Streamline and Interscope Records.

\noindent Modified Query Candidates: \\
\noindent Who released the album that takes the first place on the Billboard leaderboard in 2013?
→ Answer: Streamline and Interscope Records

\section{Implementation Details}
\label{baseline}
This section provides a detailed account of the implementation procedures for the baseline methodologies and TableRAG to ensure a fair and reproducible evaluation.

\subsection{ReAct}
For the ReAct framework, we preprocess tabular data by converting it into markdown-formatted plain text. To ensure consistency in experimental conditions, we adopt the same chunking and retrieval configurations as those employed in our TableRAG model. We build upon the publicly available ReAct implementation\footnote{\url{https://github.com/ysymyth/ReAct}}. The framework addresses user queries through an iterative reasoning loop consisting of \texttt{Thought}, \texttt{Action}, and \texttt{Observation} steps, culminating in a final answer via the \texttt{Finish} operation. The max iteration is set to 5, the same as TableRAG.

\subsection{TableGPT2}
We evaluate TableGPT2 using its officially released TableGPT Agent implementation \footnote{\url{https://github.com/tablegpt/tablegpt-agent}}. As specified in its API documentation, we provide both the document content and the tabular data as inputs to the \texttt{HumanMessage} class, ensuring adherence to the intended usage of the model:

\lstdefinestyle{mystyle}{
    commentstyle=\color{YellowGreen},
    keywordstyle=\color{magenta},
    numberstyle=\tiny\color{gray},
    stringstyle=\color{purple},
    basicstyle=\ttfamily\footnotesize,
    breakatwhitespace=false,         
    breaklines=true,                 
    captionpos=b,                    
    keepspaces=true,                 
    numbers=left,                    
    numbersep=5pt,                  
    showspaces=false,                
    showstringspaces=false,
    showtabs=false,                  
    tabsize=2
}

\lstset{style=mystyle}
\begin{lstlisting}[language=Python]
from typing import TypedDict
from langchain_core.messages import HumanMessage

class Attachment(TypedDict):
    """Contains at least one dictionary with the key filename."""
    filename: str

attachment_msg = HumanMessage(
    content="",
    # Please make sure your iPython kernel can access your filename.
    additional_kwargs={"attachments": [Attachment(filename="titanic.csv")]},
)
\end{lstlisting}

\begin{table*}[!t]
\centering
\setlength{\tabcolsep}{15pt}
\begin{tabular}{l|l|cccccccc}
\toprule[1.5pt]
\textbf{Method} & \textbf{Backbone} & \multicolumn{1}{c}{\textbf{HybridQA}} & \textbf{WikiTQ} &  \multicolumn{3}{c}{\textbf{HeteQA}} \\
\midrule
ReAct & Claude-3.5 & 41.4 & 72.6 & 19.4 \\
& DeepSeek-V3 & 35.4 & 58.4 & 15.5 \\
& Qwen-2.5-72b & 29.5 & 52.6 & 12.5  \\
\midrule
\midrule
TableGPT2 &  & 11.8 & 66.8 & 15.1 \\
\midrule
\midrule
TableRAG & Claude-3.5 & 42.6 & 86.4 & 27.8 \\
& DeepSeek-V3 & 33.1 & 79.8 & 21.7 \\
& Qwen-2.5-72b & 44.6 & 76.0 & 25.5 \\
\bottomrule[1.5pt]
\end{tabular}
\caption{Performance of TableRAG compared to baseline models across multiple benchmarks, measured by exact match.}
\label{em}
\end{table*}

\subsection{TableRAG}
In this section, we provide additional details about the offline stage in TableRAG. To mitigate the potential effects of errors during this stage, we apply various engineering techniques to minimize preprocessing mistakes:
\begin{itemize}[leftmargin=*]
    \item[-] Missing Value Imputation: we systematically assign default missing column names or missing elements to preserve table integrity.
    \item[-] Column Type Mapping: we perform data type inference for each column based on type consistency across its values.
    \item[-] Representative Sample Selection: during schema extraction, we carefully control the length of selected samples while promoting value diversity.
\end{itemize}
TableRAG is designed to degrade gracefully in the presence of offline errors (most commonly happens in the tabular processing phase). TableRAG bypasses the SQL execution pathway and leverages the text retrieval module to extract and reason over markdown-formatted table content.

\begin{table}
\centering
\begin{tabular}{l|cccccccc}
\toprule[1.5pt]
\textbf{Method} & \textbf{WikiTQ} &  \textbf{HybridQA} \\
\midrule
ReAct & 53.80 & 37.38 \\
TableRAG & 78.00 & 48.52  \\
TAP4LLM & - & 23.28 \\
H-Star & 67.35 & - \\
\bottomrule[1.5pt]
\end{tabular}
\caption{Supplementary baseline comparison with Qwen2.5-72B-Instruct as backend.}
\label{more_exp}
\end{table}

\section{Experiments}
\label{append_exp}

\subsection{Exact Match Results}
We present the experimental results with exact match as evaluation metrics in Table \ref{em}.

\subsection{Supplementary Baseline Comparison}
We further compare TableRAG with additional baseline methods that target specific domain scenarios:
\begin{itemize}[leftmargin=*]
    \item[-] TAP4LLM\footnote{\url{https://github.com/Y-Sui/archive_code}} \cite{sui2023tap4llm}: a comprehensive pre-processing toolkit designed to enhance in Tabular reasoning with LLMs. It employs three main strategies: table sampling, table augmentation and table packing.
    \item[-] H-Star\footnote{\url{https://github.com/nikhilsab/H-STAR}} \cite{abhyankar2024h}: a hybrid approach that operates in two stages: Table Extraction, which identifies and isolates relevant table sections based on the query, and Adaptive Reasoning, which dynamically selects between symbolic and semantic reasoning strategies depending on the question type.
\end{itemize}
The results are summarized in Table \ref{more_exp}. As shown, TAP4LLM performs worse than both TableRAG and ReAct, our primary baseline, indicating that agentic frameworks are generally more effective than pipeline-style approaches. In the TableQA scenario, TableRAG also surpasses H-Star. Overall, TableRAG consistently delivers strong performance, further supporting our claims.

\section{Analysis}
\label{analysis}
We present a series of auxiliary analyses conducted on both our proposed benchmark, \textsc{HeteQA}, and the publicly available \textsc{HybridQA} dataset. These analyses offer further insight into the behavior and limitations of TableRAG beyond the primary evaluation metrics.

\begin{figure}
    \centering
    \includegraphics[width=\linewidth]{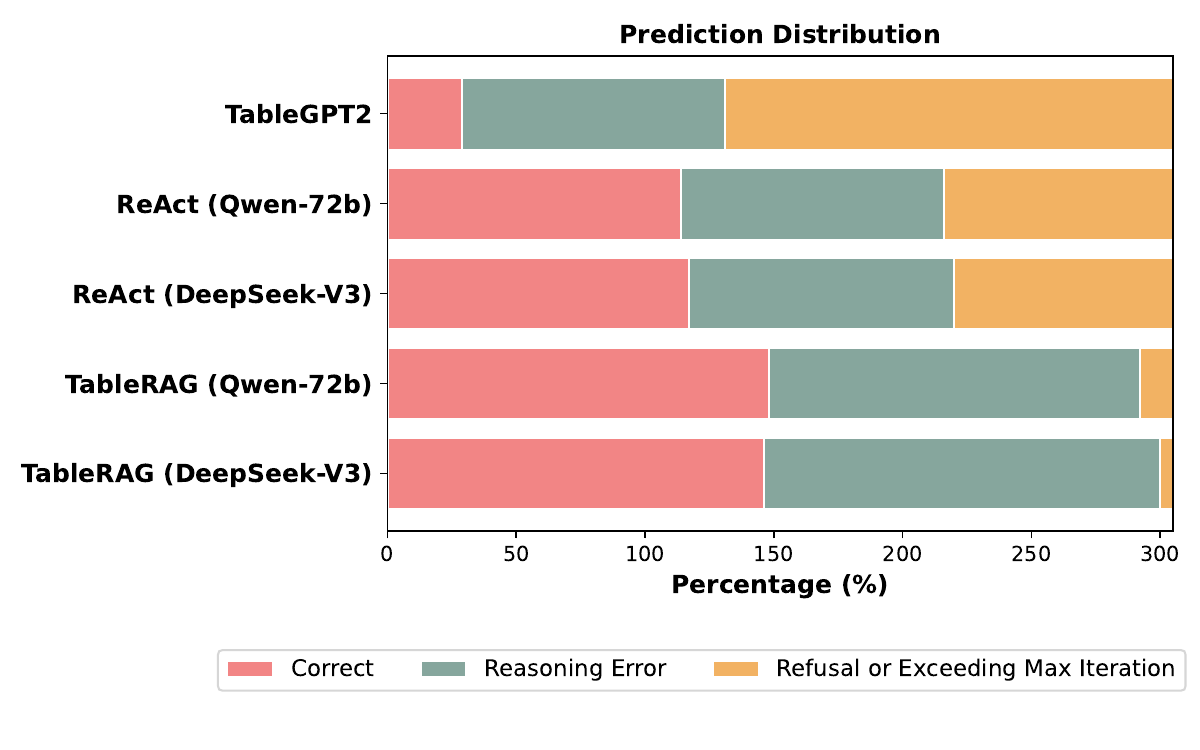}
    \caption{Prediction distribution of TableRAG, TableGPT2 and ReAct on HybridQA.}
    \label{fig:hy_failure}
\end{figure}

\begin{figure}[!t]
    \centering
    \includegraphics[width=0.6\linewidth]{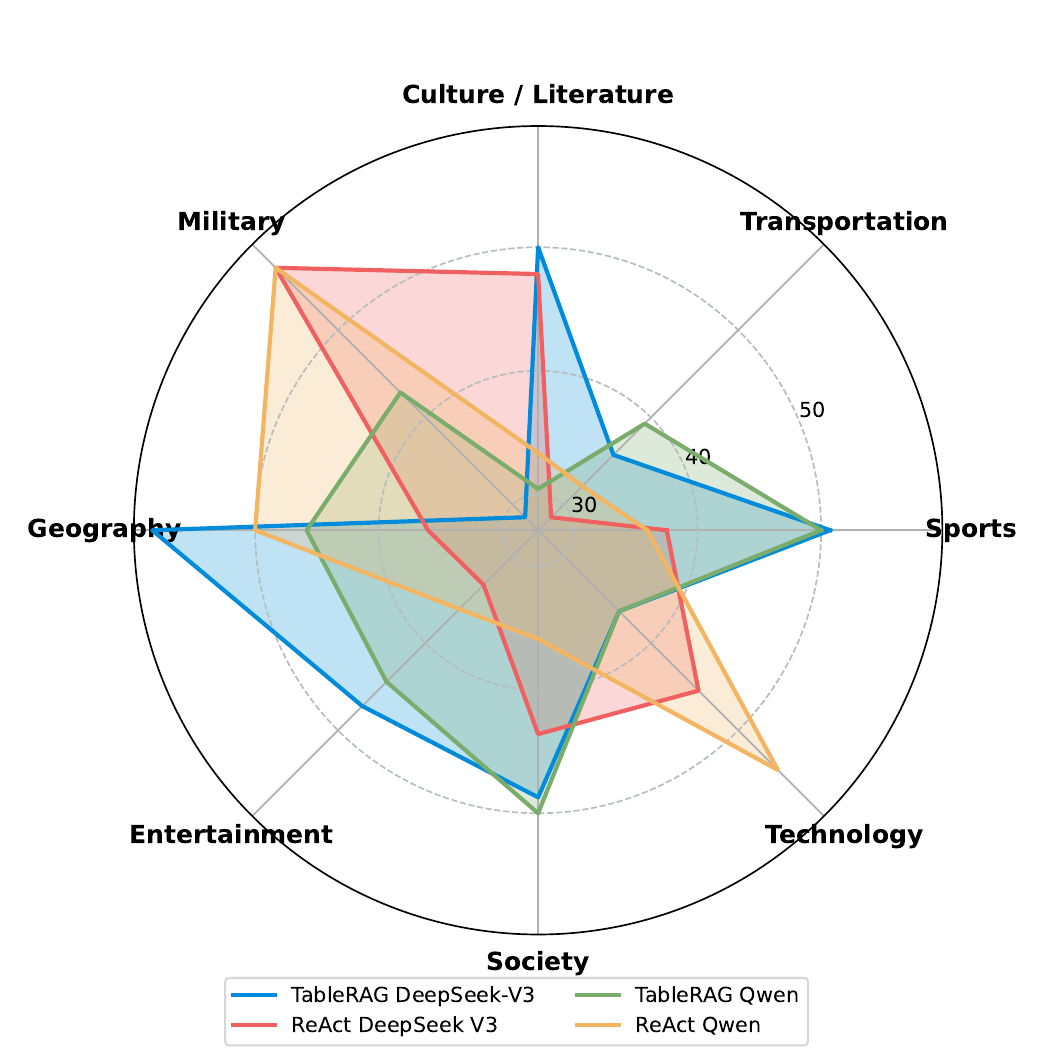}
    \caption{Model performance of different domains on HybridQA.}
    \label{fig:hy_domain}
\end{figure}

\subsection{Analysis on HybridQA}
We extend our investigation to the \textsc{HybridQA} dataset by examining the performance distribution on data domains and conducting a detailed error analysis, using \textsc{DeepSeek-v3} as the backbone model. The results, summarized in Figures~\ref{fig:hy_domain} and~\ref{fig:hy_failure}, reveal patterns consistent with those observed on \textsc{HeteQA}. This parallel further validates the generality of our observations across HeteQA settings.

\begin{figure}
    \centering
    \includegraphics[width=\linewidth]{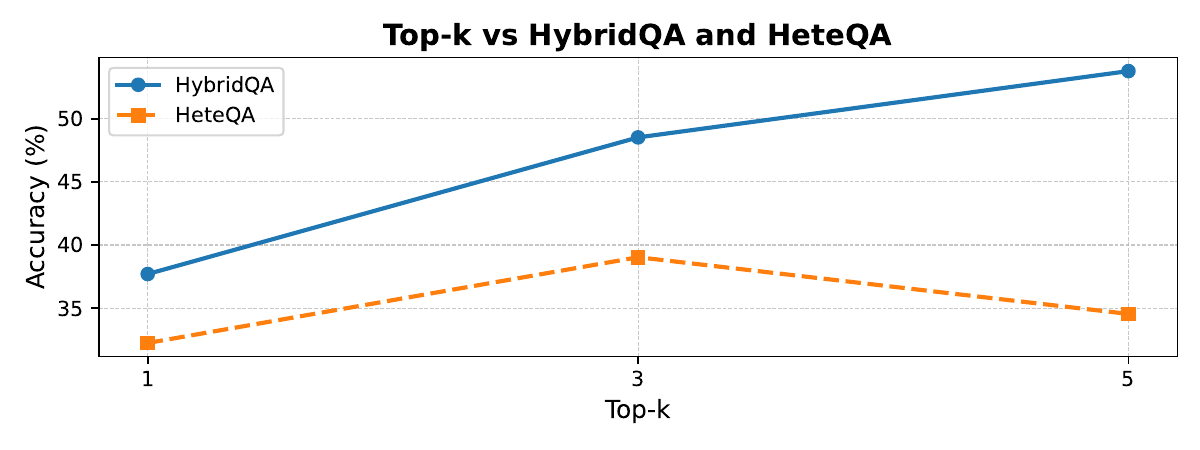}
    \caption{Hyperparameter top-$k$ analysis of TableRAG on HeteQA.}
    \label{fig:hyper}
\end{figure}

\subsection{HyperParameter Analysis}
To elucidate the influence of the top-$k$ retrieval parameter on the performance of TableRAG, we undertook a systematic sensitivity analysis. While our main experimental setup fixed $k=3$, we expanded our investigation on the HeteQA dataset, utilizing DeepSeek-V3 as the retrieval backbone, and varied $k$ across the set ${1, 3, 5}$. The resulting performance metrics are illustrated in Figure~\ref{fig:hyper}.

Our observations reveal distinct behaviors across benchmarks with respect to the choice of $k$. Notably, HybridQA exhibits superior performance at higher $k$ values. This effect is plausibly attributable to the lack of deduplication within its tables, whereby overlapping or similar information from multiple tables or Wikipedia documents contributes constructively during retrieval.
In aggregate, setting $k=3$ strikes an effective balance, yielding robust performance across both benchmarks while maintaining computational efficiency. This efficiency gain arises from the direct relationship between the top-$k$ retrieval size and the subsequent LLM context length, underscoring the practical importance of this hyperparameter in optimizing the trade-off between accuracy and resource consumption.


\section{Check List}

\paragraph{Harmful information And Privacy}
We propose a new RAG solution to address multi-hop problems related to heterogeneous data, without involving any harmful information or privacy. In addition, we provide a heterogeneous benchmark containing tables and texts sourced from Wikipedia. All of the data is publicly available, contains no personal information, and involves no harmful content.

\paragraph{License and Intend}
We provide the license we used here:
\begin{itemize}[leftmargin=*]
    \item[-] Claude 3.5 Sonnet (\url{https://www.anthropic.com/legal/aup})
    \item[-] Qwen2.5-72B-Instruct (\url{https://huggingface.co/Qwen/Qwen2.5-72B-Instruct/blob/main/LICENSE})
    \item[-] DeepSeek-V3 (\url{https://huggingface.co/deepseek-ai/DeepSeek-V3/blob/main/LICENSE-MODEL})
    \item[-] TableGPT Agent (Apache License 2.0) and TableGPT2-7B (\url{https://huggingface.co/tablegpt/TableGPT2-7B/blob/main/LICENSE})
\end{itemize}
Our use of these existing artifacts was consistent with their intended use.

\paragraph{Documentation of the artifacts}
We propose a novel Retrieval-Augmented Generation (RAG) framework that integrates traditional document retrieval with structured data querying via SQL, aiming to enhance performance in table-related question answering tasks. The framework comprises two fundamental stages: offline database construction and online interactive reasoning. Compared to conventional document-centric RAG approaches, our architecture offers additional reliable SQL execution results—when table fragments are involved in the retrieval process. This supplementary source mitigates the limitations of generative models in handling structured tabular data.

To facilitate a more comprehensive evaluation of our framework, we further construct a heterogeneous benchmark named HeteQA. This benchmark aims to evaluate the capability to handle multi-hop reasoning tasks across heterogeneous documents. The benchmark instances are initially generated by LLMs, and then rigorously validated by experienced programmers and database administrators to ensure correctness and realism.

\section{Prompts}
\label{prompts}
The prompts for TableRAG are presented in this section.
\newpage

.
\newpage

\begin{tcolorbox}[title = {Prompt Template for SQL generation}, width=\textwidth, fontupper=\small]
\# Multi-Hop Table Reasoning Query Generator \\

\#\# Task \\
Generate a genuine multi-hop reasoning query based on the provided markdown table, along with SQL and pandas solutions in a structured JSON format.\\

\#\# Input\\
A markdown formatted table schema.\\

\#\# Output Requirements\\
Provide exactly ONE multi-hop reasoning query that: \\
- Requires sequential analytical operations where each step depends on the previous result \\
- Cannot be broken down into separate independent questions \\
- Is solvable using both SQL and pandas\\
- Is relevant to the data domain in the table\\
- Mention the table name in query to indicate the source table file\\

\#\# Operations to Consider in Your Sequential Reasoning Chain\\
- Group or Aggregate \\
- Filtering subsets\\
- Calculating percentages or ratios between groups\\
- Comparing specific subgroups \\
- Rank or order\\
- Finding extremes (max/min) of aggregated values\\
- Computing difference or sum\\
- Finding correlations\\

\#\# Example Operations Combinations\\
- Filter → Group → Rank in groups\\
- Group → Sum → Compare\\
- Filter → Calculate percentage → Rank\\
- Group → Aggregate → Filter on aggregate\\

\#\# Guidance for True Multi-Hop Queries\\
A proper multi-hop query requires sequential operations where each step builds on the result of the previous step. For example:\\

GOOD (True multi-hop): "What was the average lap time among the top 5 ranked drivers in the team which had the best average lap time?"\\
 - This requires first finding the team with best average lap time\\
 - Then identifying the top 5 drivers in that specific team\\
 - Finally calculating the average lap time of just those drivers\\

 BAD (Separable questions): "Which team had the best average lap time, and what was that average among the top 5 ranked drivers?"\\
 - This could be answered as two separate questions\\

Format your response as a single JSON object with this structure:\\
\{ \\
  "query": "Clear natural language question requiring true multi-hop reasoning", \\
  "operations\_used": ["List operations used, such as: filtering, aggregation, grouping, sorting, etc."], \\
  "sql\_solution": "Complete executable SQL query that solves the question", \\
  "pandas\_solution": "Complete executable pandas code that solves the question", \\
  "result\_type: "The type of the result, must be either number or entity." \\
\} 
\\
Ensure your query truly requires chained reasoning where later steps must use results from earlier steps and generates one answer.
\end{tcolorbox}

\newpage
.
\newpage

\begin{tcolorbox}[title = {Prompt Template for Query Decomposition}, width=\textwidth]
\begin{lstlisting}
tools = [{
    "type": "function",
    "function": { 
        "name": "solve_subquery", 
        "description": "Return answer for the decomposed subquery",
        "parameters": { 
            "type": "object", 
            "properties": { 
                "subquery": { 
                    "type": "string", 
                    "description": "The subquery to be solved" 
                } 
            }, 
            "required": [ 
                "subquery" 
            ], 
            "additionalProperties": False 
        }, 
        "strict": True 
    } 
}] 
\end{lstlisting}
Next, You will complete a table-related question answering task. Based on the provided materials such as the table content (in Markdown format), you need to analyze the Question. And try to decide whether the Question should be broken down into subquerys. After you have collected sufficient information, you need to generate comprehensive answers. \\
You have a "solve\_subquery" tool that can execute SQL-like operations on the table data. It accepts natural language questions as input. \\

Instructions: \\
1. Carefully analyze the user query through step-by-step reasoning. \\
2. If the query requires multiple pieces of information, more than the given table content: \\
   - Decompose the query into subqueries \\
   - Process one subquery at a time \\
   - Use "solve\_subquery" tool to retrieve answers for each subquery \\
3. If a query can be answered by table content, do not decompose it. And directly put the origin query into the "solve\_subquery" tool. \\
   The "solve\_subquery" tool can solve complex subquery on table via one tool call. \\
4. Generate exactly ONE subquery at a time. \\
5. Write out all terms completely - avoid using abbreviations. \\
6. When you have sufficient information, provide the final answer in this format: \\
   <Answer>: [your complete response] \\
   
Table Content: \{table\_content\} \\
Question: \{query\} \\
Please start! \\
\end{tcolorbox}

\newpage
???
\newpage

\begin{tcolorbox}[title = {Prompt Template for Intermediate Answer Reasoning}, width=\textwidth]
You are about to complete a table-based question answering task using the following two types of reference materials: \\

\# Content 1: Original content (table content is provided in Markdown format) \\
\{docs\} \\

\# Content 2: NL2SQL related information and SQL execution results in the database \\
\# the user given table schema \\
\{schema\} \\

\# SQL generated based on the schema and the user question: \\
\{nl2sql\_model\_response\} \\

\# SQL execution results \\
\{sql\_execute\_result\} \\

Please answer the user's question based on the materials above. \\
User question: \{query\} \\

Note: \\
1. The markdown table content in Content 1 may be not complete. \\
2. You should cross-validate the given two materials:\\
   - if the answers are same, you may directly output the answer.\\
   - If the SQL shows error, such as "SQL execution results", try to answer solely based on Content 1.\\
   - If the two material shows conflict, carefully evaluate both sources, explain the discrepancy, and provide your best assessment. \\

\end{tcolorbox}

\newpage
.
\newpage

\begin{tcolorbox}[title = {Prompt Template for Answer Evaluation}, width=\textwidth]
We would like to request your feedback on the performance of the AI assistant in response to the user question displayed above according to the gold answer. Please use the following listed aspects and their descriptions as evaluation criteria: \\
    - Accuracy and Hallucinations: The assistant's answer is semantically consistent with the gold answer; The numerical value and order need to be accurate, and there should be no hallucinations. \\
    - Completeness: Referring to the reference answers, the assistant's answer should contain all the key points needed to answer the user's question; further elaboration on these key points can be omitted. \\
Please rate whether this answer is suitable for the question. Please note that the gold answer can be considered as a correct answer to the question. \\

The assistant receives an overall score on a scale of 0 OR 1, where 0 means wrong and 1 means correct. \\
Dirctly output a line indicating the score of the Assistant. \\

PLEASE OUTPUT WITH THE FOLLOWING FORMAT, WHERE THE SCORE IS 0 OR 1 BY STRICTLY FOLLOWING THIS FORMAT: "[[score]]", FOR EXAMPLE "Rating: [[1]]": \\
<start output> \\
Rating: [[score]] \\
<end output> \\

[Question] \\
{question} \\

[Gold Answer] \\
{golden} \\

[The Start of Assistant's Predicted Answer] \\
\{gen\} \\

\end{tcolorbox}

\end{document}